\pdfoutput=1

\documentclass[11pt]{article}

\usepackage[final]{coling}

\usepackage{times}
\usepackage{latexsym}

\usepackage[T1]{fontenc}

\usepackage[utf8]{inputenc}

\usepackage{microtype}

\usepackage{inconsolata}

\usepackage{graphicx}
\usepackage{adjustbox}
\title{From Chain to Tree: Refining Chain-like Rules into Tree-like Rules on Knowledge Graphs}

\author{
  Wangtao Sun$^{1,2}$, Shizhu He$^{1,2}$\thanks{corresponding author}, Jun Zhao$^{1,2}$, Kang Liu$^{1,2,3}$\footnotemark[1] \\
  \textit{$^{1}$Institute of Automation, Chinese Academy of Sciences, Beijing, China} \\
  \textit{$^{2}$School of Artificial Intelligence, University of Chinese Academy of Sciences, Beijing, China} \\
  \textit{$^{3}$Shanghai Artificial Intelligence Laboratory} \\
}

\usepackage{microtype}

\usepackage{graphicx}
\usepackage{tabularx}
\usepackage{soul}

\usepackage{amsmath,bm}
\usepackage{amssymb}
\usepackage{mathrsfs}
\usepackage{amsthm,amsmath,amssymb}
\usepackage{mathrsfs}
\usepackage{algpseudocode}
\usepackage{graphicx}
\usepackage{epstopdf}
\usepackage{tcolorbox}
\usepackage{latexsym}
\usepackage{arydshln}
\usepackage{subfigure}
\usepackage{xcolor}
\usepackage{booktabs}
\usepackage{multirow}
\usepackage{enumitem}

\definecolor{myblue}{RGB}{0,176,240} 
\definecolor{mygreen}{RGB}{0,176,80} 
\definecolor{mypurple}{RGB}{112,48,160} 

\begin{document}
\maketitle


\begin{abstract}
With good explainability and controllability, rule-based methods play an important role in the task of Knowledge Graph Completion (KGC). However, existing studies primarily focused on learning chain-like rules, whose chain-like structure limits their expressive power. Consequently, 
chain-like rules often exhibit lower Standard Confidence, and are prone to the incorrect grounding values during reasoning, thus producing erroneous reasoning results.
In this paper, we propose the concept of tree-like rules on knowledge graphs to expand the scope of the application and improve the reasoning ability of rule-based methods. To achieve this, we formalize the problem of tree-like rule refinement and propose an effective framework for refining chain-like rules into tree-like rules. 
Experimental evaluations on four public datasets demonstrate that the proposed framework can seamlessly adapt to various chain-like rule induction methods and the refined tree-like rules consistently exhibit higher Standard Confidence and achieve better performances than the original chain-like rules on link prediction tasks. Furthermore, we illustrate that the improvements brought by tree-like rules are positively correlated with the density of the knowledge graphs. The data and code of this paper can be available at https://github.com/forangel2014/tree-rule.
\end{abstract}

\section{Introduction}

\noindent
Knowledge Graph Completion (KGC)~\cite{taskar2003link,10031226,wang2023survey} is a fundamental and important task in Natural Language Processing. For KGC, rule-based methods play a pivotal role, which focus on first learning symbolic and interpretable rules, and then leverage them for effective reasoning within Knowledge Graphs (KGs). In these rule-based methods, the development of comprehensive and high-quality rule sets is essential for the success of KGC~\cite{nandi2023simple, anyburl}.

\begin{figure}[t]
\begin{center}    
\begin{adjustbox}{max width=\columnwidth}    
    \includegraphics[width=0.5\textwidth]{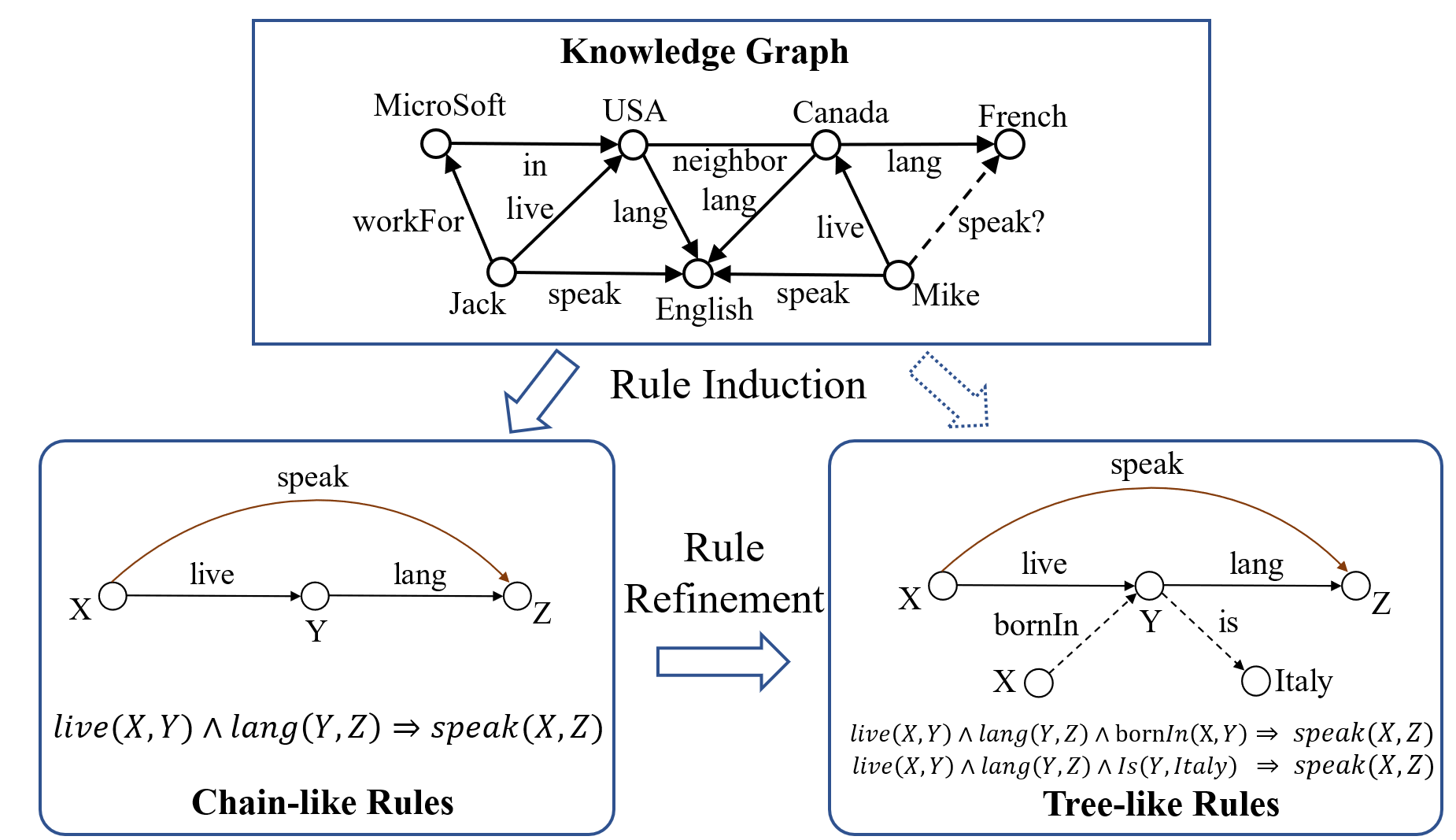}
\end{adjustbox}
\caption{An example of an inaccurate chain-like rule and the refined tree-like rule. Although the chain-like rule (bottom-left) can predict most cases correctly, due to its chain structure, it has limited expressive power. The refined tree-like rule (bottom-right) leverages the information in the KG that originally ignored to improve the chain-like rule.}
\label{figure1}
\end{center}
\end{figure}

To this end, previous works have proposed various types of methods to induce rules from the KGs, like symbol-based \cite{amie}, embedding-based \cite{RLvLR, rnnlogic, ncrl}, Differentiable-ILP-based \cite{neuralLP, NLIL}, and RL-based \cite{anyburl}. 
However, these existing rule induction methods only consider the chain-like rules in KGs \cite{amie, neuralLP, RLvLR, anyburl, rnnlogic, ncrl}.
chain-like rules are a special case of the Horn Clauses \cite{aiamodernapproach}, which is equivalent to a multi-hop reasoning process \cite{NLIL}. For example, in the left-bottom subfigure in Figure \ref{figure1}, $\textrm{live}(X, Y) \wedge \textrm{lang}(Y, Z) \Rightarrow \textrm{speak}(X, Z)$ is a typical chain-like rule. Since there is only one path leading the query variable $X$ to the target variable $Z$, there are limited constraints for the semantics in the reasoning path of the chain-like rules. Consequently, chain-like rules are often prone to the incorrect grounding values during reasoning, thus producing erroneous reasoning results. For example, in Figure \ref{figure1}, this rule may reason erroneous facts when $X$ is just a \textit{traveler}, or $Y$ is a \textit{country} with more than one official language and $X$ just \textit{speaks} one of them.

Therefore, this paper introduces the concept of \emph{tree-like rules}, which is a more general form of rules. Apart from a direct path from the query variable $X$ to the target variable $Z$, the tree-like rule body also contains some branch atoms to constrain the grounding values of the rule. These branch atoms can be seen as ``hanging'' triplets on the path, which further shape and narrow down the possible grounding values of the rule. The path and the branch atoms stretched together into a tree structure in the KG. For example, in the right-bottom subfigure in Figure \ref{figure1}, the tree-like rule may look like $\textrm{live}(X, Y) \wedge \textrm{lang}(Y, Z) \wedge \textrm{bornIn}(X, Y) \Rightarrow \textrm{speak}(X, Z)$ or $\textrm{live}(X, Y) \wedge \textrm{lang}(Y, Z) \wedge \textrm{is}(Y, \textrm{Italy}) \Rightarrow \textrm{speak}(X, Z)$. The added atoms $\textrm{bornIn}(X, Y)$ and $\textrm{is}(Y, \textrm{Italy})$ yield new constraints for the grounding values of variable $Y$. With these added constraints, tree-like rules are expected to possess higher quality (Standard Confidence) and avoid the wrong predictions that \emph{X speaks Z} in the cases mentioned before. As a result, tree-like rules possess higher Standard Confidence than chain-like rules and will exhibit better reasoning performance on the task of KGC.

To refine chain-like rules into tree-like rules, the key challenge is to tackle the large combinatorial space of the rule body, i.e. searching and selecting the probable branch atoms. The branch atoms are supposed to exclude the incorrect groundings while still including the correct groundings of the rule. To refine chain-like rules into tree-like rules by adding branch atoms, our proposed framework first transforms the optimization problem of the Standard Confidence of the entire rule to that of the \textbf{best branch atom selection} in a specific variable in the rule body. 
For the best branch atom selection problem, our framework proposes a three-step pipeline: Forward Reasoning, Backward Reasoning, and Candidate Atom Selections. Through this pipeline, our framework effectively refines the original chain-like rules into high-quality tree-like rules.

\begin{figure*}[t]
\begin{center}
\includegraphics[width=0.85\textwidth]{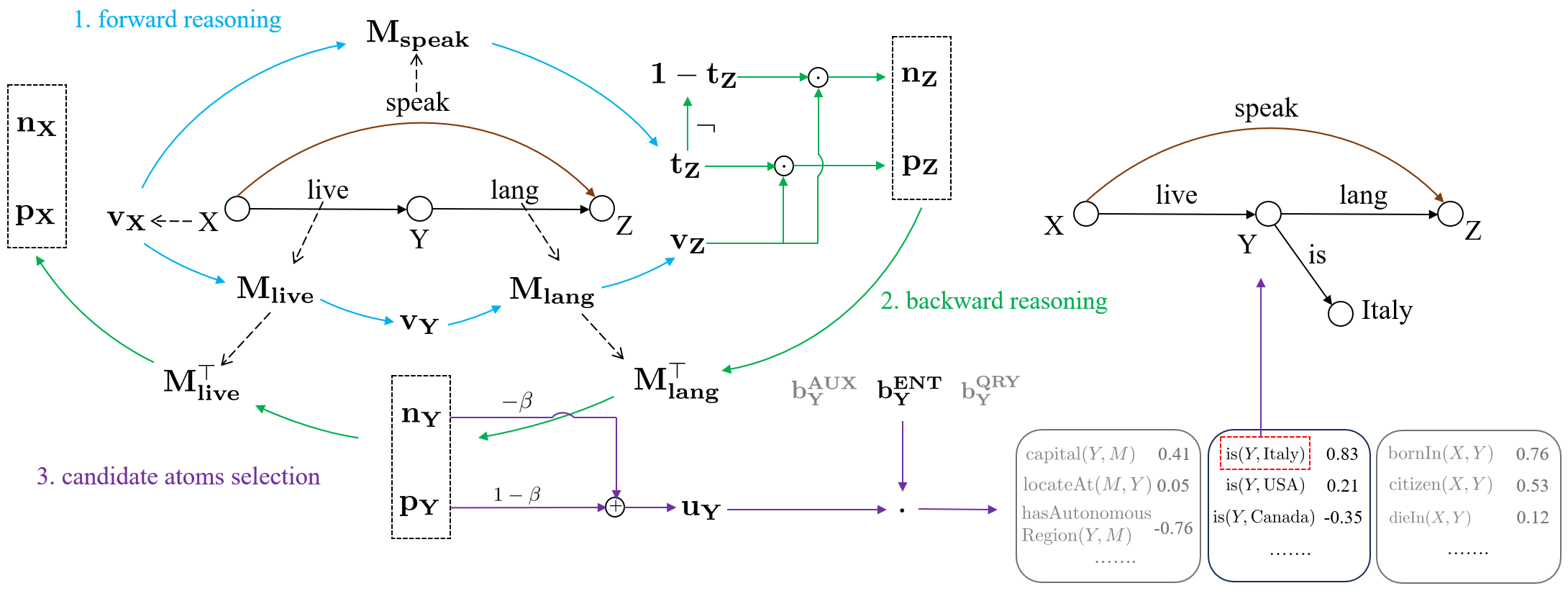}
\caption{The framework of our proposed method. In the \textcolor{myblue}{Forward Reasoning} stage, the Query variable $X$ is first grounded with $b$ randomly sampled entities and by forward reasoning, we obtain the grounding values of $Y$ and $Z$. In the \textcolor{mygreen}{Backward Reasoning} stage, we then abductively obtain the positive groundings and negative groundings of each variable in the rule body. Finally in the \textcolor{mypurple}{Candidate Atoms Selection} stage, three types of candidate branch atoms are then selected according to their inner product scores with the variable representation.}
\label{figure2}
\end{center}
\end{figure*}



To verify the effectiveness of tree-like rules refined by our framework, we conduct experiments on four widely used benchmark KGs with three different sources of chain-like rules. 
The experimental results show that tree-like rules continuously exhibit higher Standard Confidence, and outperform chain-like rules on the link prediction tasks for different sources of chain-like rules on different KGs. With further analysis, we also find that the improvements brought by tree-like rules are positively correlated with the density of the knowledge graphs, showing that tree-like rules have greater advantages in KGs with complex topological structures.

In summary, the contributions of this paper are as follows:
\begin{itemize}[itemsep=1pt,topsep=1pt,parsep=0pt,leftmargin=*]
    \item This paper proposes the concept of tree-like rules for the task of rule induction on KGs. An effective framework is proposed for refining chain-like rules induced from any existing method into tree-like rules. 
    \item The paper conducts experiments on four open-accessed datasets and the results show our refined tree-like rules from different chain-like rules consistently have higher Standard Confidence, and outperform on KG reasoning task than original chain-like rules. Further analysis finds that the improvements brought by tree-like rules are positively correlated with the density of the knowledge graphs.
\end{itemize}  

\section{Problem Formulation}
\label{sec:formulation}
\noindent
In the scope of First-Order Logic (FOL), the rule (or Horn Clause) is formalized as $\sigma \Rightarrow \varphi$. Here, the left part $\sigma$ is called ``rule body'', which serves as the premise, when it is satisfied by some groundings, then the right part, ``rule head'' $\varphi$, will be grounded as the conclusion. 

For the evaluation of the quality of such a rule, as we are adding constraints to refine it to be more precise,
we adopt the widely-used metric, Standard Confidence (sc). It can be defined as:
\begin{equation}
    sc = \frac{\# (S_{\sigma} \cap S_{\varphi})}{\# S_{\sigma}}
\label{eq:base_objective}
\end{equation}
where $\#(\cdot)$ stands for the count of possible groundings. This metric can also be easily understood if we take the \emph{rule} as a binary classifier in machine learning: $S_{\sigma}$ stands for the situation that the classifier output ``positive'', and $S_{\varphi}$ stands for the situations that are ``true''. Therefore, \emph{sc} corresponds to the \emph{precision}, being a key metric to describe how much we can trust the rule.

Based on the definition and objective above, our task can be stated as follows. A given chain-like rule $\mathscr{R}$ of length $n$ can be represented as:
\begin{equation}
    \begin{aligned}
    \mathscr{R} : r_0(x_0, x_1) \wedge ...
    \wedge r_{n-1}(x_{n-1}, x_n) \Rightarrow r(x_0, x_n) \nonumber
    \end{aligned}
\end{equation}
\noindent

To obtain tree-like rules from it, we aim to find branch atoms $b(x_i)$ for the variable $x_i$ while being aware of the objective in Eq~\ref{eq:base_objective}. Finally, a refined tree-like rule has the following format:
\begin{equation}
    \begin{aligned}
    \mathscr{R}^{*} : &r_0(x_0, x_1) \wedge ... \wedge r_{i-1}(x_{i-1}, x_i) \wedge b(x_i) \\
    &\wedge ... \wedge r_{n-1}(x_{n-1}, x_n) \Rightarrow r(x_0, x_n) \nonumber
    \end{aligned}
\end{equation}
Please note that when we represent branch atoms as $b(x_i)$, we omit other variables and constants that may appear within the predicate, to emphasize that this branch is a constraint on the variable $x_i$.

\section{Method}
\label{method}
\noindent
Given the chain-like rule, this paper propose to first grounded the rules with entities in the KG, and then find the top branch atoms that could eliminate the undesirable groundings, to make the groundings satisfying rule body be as close as to those satisfying rule head.

Given a KG $\mathcal{G} = (\mathcal{E}, \mathcal{R}, \mathcal{T})$, where $\mathcal{E}, \mathcal{R}, \mathcal{T}$ stands for the entity set, relation set, and triplet set, respectively, to reason with the rule on this KG, this paper adopts the matrix representation of entities and relations for reasoning rules. 

For an entity $e \in \mathcal{E}$, let $\mathbf{v_e} \in \{0,1\}^{1 \times \vert \mathcal{E} \vert}$ be the one-hot encoding of entity $e$, i.e. only the $i$th element is 1 if $e$ is the $i$th entity in $\mathcal{E}$. 
Based on the encoding of single entity, for a variable $x$ in the rule, if the set $C_x$ denote the entities that can be ground to $x$, then the variable grounding of $x$, $\mathbf{v_x} \in \{0,1\}^{1 \times \vert \mathcal{E} \vert}$, is defined as $\mathbf{v_x} = \sum_{e \in C_x} \mathbf{v_e}$.

For an relation $r \in \mathcal{R}$, let $\mathbf{M_r}\in \{0,1\}^{\vert \mathcal{E} \vert \times \vert \mathcal{E} \vert}$ be the binary adjacency matrix of relation $r$, i.e. $\mathbf{M_r}[i,j] = 1$ iff $(e_i, r, e_j) \in \mathcal{T}$. 
With the definition above, a reasoning hop $(h,r,t?)$ can be modeled by the following matrix multiplication:
\begin{equation}
    \begin{aligned}
        \mathbf{v_{t}} = \mathbf{v_{h}} \mathbf{M_{r}}
    \end{aligned}
\label{eq:base_reasoning}
\end{equation}

To obtain $\mathscr{R}^*$ from $\mathscr{R}$, we need to find a branch atom $b(x_i)$ (correspond to a constraint vector $\mathbf{b_{x_i}}$) for the ground value of variable $x_i$, to best match the reasoning results between rule body and rule head, based on Eq~\ref{eq:base_objective}, we have:
\begin{equation}
\small
    \begin{aligned}
        J_{\mathscr{R}} =& \left\| \mathbf{v_{x_0}} \mathbf{M_{r}} \odot 
        [( \mathbf{v_{x_0}} \prod \limits_{j=0}^{i-1} \mathbf{M_{r_j}} \odot \mathbf{b_{x_i}} ) \prod \limits_{j=i}^{n-1}\mathbf{M_{r_j}}] \right\| \\
        &/ \left\| (\mathbf{v_{x_0}} \prod \limits_{j=0}^{i-1} \mathbf{M_{r_j}} \odot \mathbf{b_{x_i}} ) \prod \limits_{j=i}^{n-1}\mathbf{M_{r_j}} \right\|
    \end{aligned}
\label{eq:expanded_objective}
\end{equation}
\noindent
where $\left\|\mathbf{v}\right\|$ stands for the 1-norm of the vector $\mathbf{v}$.

However, it is hard to directly find branch atoms by maximizing Eq~\ref{eq:expanded_objective}. Thus we approximately transform it into the best branch atom selection problem:
\begin{equation}
\small
    \begin{aligned}
        J_{\mathscr{R}}^{'} = S \left( 
        \textcolor{myblue}{\mathbf{v_{x_0}} \mathbf{M_{r}}} 
        \textcolor{mygreen}{\prod \limits_{j=n-1}^{i}{\mathbf{M_{r_j}}^\top}},
        \textcolor{myblue}{\mathbf{v_{x_0}} \prod \limits_{j=0}^{i-1} \mathbf{M_{r_j}}} 
        \textcolor{mypurple}{\odot \mathbf{b_{x_i}}} \right)
    \end{aligned}
    \label{eq:final_objective}
\end{equation}
\noindent
where $S(\mathbf{a},\mathbf{b}) = (1-\beta)\left\|\mathbf{a} \odot \mathbf{b} \right\| - \beta \left\| (1-\mathbf{a}) \odot \mathbf{b} \right \|$ is a similarity metric balancing the excluding of the incorrect groundings and the including of the correct groundings. Here, the $\odot$ denotes the element-wise product, and $\beta \in (0,1)$ balances the degree of including positive groundings and excluding negative groundings. The colors in this equation correspond to the colors in Figure~\ref{figure2}.

To find the best branch atoms to maximize Eq~\ref{eq:final_objective}, as shown in Figure~\ref{figure2}, we propose a three-step framework:
\begin{itemize}[itemsep=1pt,topsep=1pt,parsep=0pt,leftmargin=*]
    \item \S\ref{forward reasoning} \textcolor{myblue}{Forward Reasoning}: we first sample a batch of $b$ entities to ground the query variable $x_0$ (i.e. $X$ in Figure~\ref{figure2}).  
    Then a forward reasoning process transfers the groundings of $x_0$ to the target variable $x_n$ (i.e. $Z$ in Figure~\ref{figure2}) through both the rule body and rule head.
    At target variable $x_n$, the positive groundings ($\mathbf{p_{x_n}}$, the entities correctly predicted by rule body) and 
    negative groundings ($\mathbf{n_{x_n}}$, the entities incorrectly predicted by the rule body) are obtained. 
    \item \S\ref{backward reasoning} \textcolor{mygreen}{Backward Reasoning}: we then abductively obtain the positive groundings and negative groundings of each variable in the rule body, by sequentially multiplying the transpose of the relation matrix with the current grounding vectors.
    \item  \S\ref{Candidates} \textcolor{mypurple}{Candidate Atoms Selection}: the variable representation is a weighted sum of its positive groundings and negative groundings, 
    which is a trade-off of including positive groundings and excluding negative groundings when adding branch atoms.
    Finally, we consider three types of branch atoms that are to be added to the rule bodies, 
    the candidate branch atoms are then selected according to their inner product scores with the variable representation.
\end{itemize}

\subsection{Forward Reasoning}
\noindent
\label{forward reasoning}
Let us first sample a batch of $b$ entities to ground $x_0$, and we concatenate their encoding to get the initial variable grounding $\mathbf{v_{x_0}} \in \{0,1\}^{b \times \vert \mathcal{E} \vert}$.
As most rules only cover a small part of entities on the entire KG, directly sample $b$ entities from the entity set $\mathcal{E}$ may involve many ``inactive tracks'' (line of all 0s in matrix) to the reasoning process, we choose to sample from the entities that at least satisfy the first relation $r_0$. This can be done by sum up the columns of $\mathbf{M_{r_0}}$ to find the candidate entities:
\begin{equation}
    \begin{aligned}
        \mathbf{v_{cand}} = \sum_j \mathbf{M}^{i,j}_{\mathbf{r_0}}
    \end{aligned}
\end{equation}

Then, by randomly keeping $b$ entities from $\mathbf{v_{cand}}$ and concatenating their one-hot encoding, we can obtain the initial variable grounding $\mathbf{v_{x_0}}$. Based on the initial variable grounding $\mathbf{v_{x_0}}$, the forward process of rule body can be modeled as a series of matrix multiplication: 

\begin{equation}
    \begin{aligned}
        \mathbf{v_{x_{i+1}}} = \mathbf{v_{x_i}} \mathbf{M_{r_i}}, i = 0,1,...,n-1
    \end{aligned}
\end{equation}
where $\mathbf{v_{x_i}}$ is the variable grounding of $x_i$, and finally the reasoning result of this rule is $\mathbf{v_{x_n}}$. 
Similarly, we can obtain the true grounding of $x_n$ by applying rule head: $\mathbf{t_{x_{n}}} = \mathbf{v_{x_0}} \mathbf{M_{r}}$. The $\mathbf{v_{x_n}}$ and $\mathbf{t_{x_n}}$ are vectorized groundings of $S_{\sigma}$ and $S_{\varphi}$ introduced in Section~\ref{sec:formulation}.


\subsection{Backward Reasoning}
\label{backward reasoning}
\noindent
After obtaining the reasoning results from both the rule body ($\mathbf{v_{x_{n}}}$) and rule head ($\mathbf{t_{x_{n}}}$) of $x_n$, we then discriminate the positive (the groundings that thought to ground the $x_n$ by rule body, and proved to satisfy the rule head as well) and negative (the groundings that thought to ground the $x_n$ by rule body, but proved not to satisfy the rule head) groundings by performing element-wise production: 
\begin{equation}
\begin{aligned}
    \mathbf{p_{x_{n}}} &= \mathbf{v_{x_n}} \odot \mathbf{t_{x_{n}}} \\
    \mathbf{n_{x_{n}}} &= \mathbf{v_{x_n}} \odot (1-\mathbf{t_{x_{n}}})
\end{aligned}    
\end{equation}
where the i-th element in $\mathbf{p_{x_{n}}}$/$\mathbf{n_{x_{n}}}$ represents how many entity $e_i$ appear as positive/negative result in rule $\mathscr{R}$. 

Notice that the encoding of $r^{-1}$ (inverse relation of $r$) is $\mathbf{M_r}^\top$. 
Then we can abductively obtain the positive and negative groundings at all variables in rule $\mathscr{R}$ by backward reasoning:

\begin{equation}
    \begin{aligned}
\mathbf{p_{x_{i}}} &= \left(\mathbf{p_{x_{i+1}}} \mathbf{M_{r_{i+1}}}^\top \right) \odot \mathbf{v_{x_{i}}} \\
\mathbf{n_{x_{i}}} &= \left(\mathbf{n_{x_{i+1}}} \mathbf{M_{r_{i+1}}}^\top \right) \odot \mathbf{v_{x_{i}}}        
    \end{aligned}
\end{equation}
By performing such a backward reasoning process, we can have the knowledge that at each variable of the rule, which entities are desired (positive) and which are undesired (negative).

\subsection{Candidate Atoms Selection}
\label{Candidates}
\noindent
After obtaining the positive and negative groundings of each variable of the rule,
we then propose to refine the chain-like rules by evaluating the candidate branches on each variable.
The branch atoms we add at a specific variable of rule $\mathscr{R}$ aims to include positive groundings while excluding the negative ones as much as possible. 
So we define the representation of each variable by linearly combining $\mathbf{p_{x_{i}}}$ and $\mathbf{n_{x_{i}}}$: 
\begin{equation}
\mathbf{u_{x_{i}}} = (1-\beta) \cdot \mathbf{p_{x_{i}}} - \beta \cdot \mathbf{n_{x_{i}}}     
\end{equation}
where $\beta \in (0,1)$ is a weight balance the degree of including positive groundings and excluding negative groundings. 
$\beta \rightarrow 0$ tend to include all positive groundings, while $\beta \rightarrow 1$ tend to exclude all negative groundings.

Now that we have obtained the vectorized representation of each variable in the rule body, we thus consider applying a binary mask $\mathbf{b_{x_i}}$ (the constraint brought by branch atom $b(x_i)$) to the $\mathbf{u_{x_{i}}}$. 
This paper considers three types of branch atoms that are to be added to the rule body.
As shown in Figure~\ref{figure2}, suppose that we add branches to constrain the groundings of variable $Y$ in the rule body:
\begin{itemize}[itemsep=1pt,topsep=1pt,parsep=0pt,leftmargin=*]
    \item \textbf{AUX}. This type of branch atom yields a one-hop result from an auxiliary variable $M$, constraining $Y$ to the entities that satisfy a certain relation. i.e. $b(Y) \Leftrightarrow r(M, Y)$. In this way, $\mathbf{b_Y} = \mathbf{1}^\top \mathbf{M_{r}}$, where $\mathbf{1}$ denotes a $\vert \mathcal{E} \vert \times 1$ column vector with all 1, which corresponds to the auxiliary $M$. For example, $b(Y) \Leftrightarrow \textrm{capital}(Y, M)$.
    \item \textbf{ENT}. This type of branch atom grounds a variable in the rule body to a unique entity. i.e. $b(Y) \Leftrightarrow \textrm{is}(e, Y)$. In this way, $\mathbf{b_Y} = \textrm{one-hot}(e)$. For example, $b(Y) \Leftrightarrow \textrm{Is}(Y, \textrm{Italy})$.
    \item \textbf{QRY}. This type of branch atom yields a one-hop result from the query variable $X$ (i.e. $x_0$ in \S\ref{sec:formulation}), constraining $Y$ to the entities that have an additional relation to $X$. i.e. $b(Y) \Leftrightarrow r(X, Y)$. In this way, $\mathbf{b_{Y}} = \mathbf{v_{X}}\mathbf{M_{r}}$. For example, $b(Y) \Leftrightarrow \textrm{bornIn}(X, Y)$.
\end{itemize}
For each KG, the candidate constraint vectors of $\textbf{AUX}$ and $\textbf{ENT}$ can be obtained through preprocessing and stay fixed during the whole refinement process. For each rule, we can obtain the constraint vectors of $\textbf{QRY}$ after the $b$ initial groundings are sampled. 

After obtaining the variable representations and constraint vectors of candidate atoms, the score of adding a branch atom $b(x_i)$ is defined by the inner product: 
\begin{equation}
    \textrm{score}(b(x_i)) = \mathbf{u_{x_{i}}} \mathbf{b_{x_i}}^\top
\end{equation} 
In the implementation, $\mathbf{u_{x_{i}}}$ is multiplied with each type of candidate atoms and we greedily select the branch atoms with the top $k$ scores for each variable appearing in the rule body.

\begin{table*}[t]
\centering
    \begin{tabular}{lllllll} 
        \toprule
        Avg. sc & \multicolumn{3}{c}{FB15k-237}  &  \multicolumn{3}{c}{WN18RR}                              \\
        Rule                 & BBFS & AMIE  & AnyBurl   & BBFS & AMIE  & AnyBurl                           \\ 
        \midrule
        CHAIN                & 12.85 & 30.71 & 26.84     & ~~5.93 & 28.05 & ~~8.66 \\
        TREE(AUX)              & 21.24  & 41.42 & 27.38  & 15.85 & 42.12 & ~~~- \\
        TREE(ENT)              & 61.34  & 76.47 & 61.12  & 88.14 & 93.63 & 87.30 \\
        TREE(QRY)              & 40.88 & 56.66 & 47.57   & 61.71 & 59.98 & 62.88 \\
        TREE              & 35.96 & 56.63 & 43.04        & 55.06 & 73.64 & 85.43 \\
        \midrule
        Avg. sc & \multicolumn{3}{c}{UMLS}  &  \multicolumn{3}{c}{YAGO3-10}  \\
                Rule                 & BBFS & AMIE  & AnyBurl   & BBFS & AMIE  & AnyBurl                           \\ 
        \midrule
        CHAIN                & 14.79 & 19.57 & 16.08     & ~~8.22 & 19.76 & 16.01 \\
        TREE(AUX)              & 19.86 & 33.29 & 26.05  & 15.56 & 27.60 & 19.09 \\
        TREE(ENT)              & 32.00 & 43.92 & 37.86  & 65.26 & 71.24 & 64.86 \\
        TREE(QRY)              & 25.65 & 39.51 & 32.19   & 41.08 & 44.48  & 47.23 \\
        TREE              & 24.48 & 38.24 & 30.93        & 37.17 & 49.18 & 42.38 \\
        \bottomrule
    \end{tabular}
    \caption{The average Standard Confidence of different rules on FB15k-237. CHAIN denotes the original chain-like rules mined by each method. AUX, ENT, and QRY denote the three kinds of branch atoms in \S\ref{Candidates}. TREE denotes the refined tree-like rules. - denotes there are no such type of rules refined.}
    \label{tab:sc}
\end{table*}

\section{Experiments}
\noindent
In this section, we conduct a series of experiments to evaluate and compare the refined tree-like rules with original chain-like rules in the following two aspects: 
\begin{itemize}[itemsep=1pt,topsep=1pt,parsep=0pt,leftmargin=*]
\item \textbf{Standard Confidence} (\S\ref{sec:sc}). We adopt the Standard Confidence as the direct metric to evaluate if the refined tree-like rules have better ``quality'' than the original chain-like rules.
\item \textbf{Link Prediction} (\S\ref{sec:lp}). To further verify the effectiveness of the refined tree-like rules, we compare two types of rules on the task of Link Prediction and validate if tree-like rules conduct better reasoning than chain-like rules.
\end{itemize}

\begin{table*}[t]
\begin{center}
\begin{tabular}{lllllllllll} 
    \hline
    \multicolumn{1}{l}{Dataset}          & \multicolumn{5}{c}{FB15k-237}                       & \multicolumn{5}{c}{WN18RR}          \\ 
    \hline
    \multicolumn{2}{l}{Rule}             & MRR            & Hit@1          & Hit@3          & Hit@10            & MRR            & Hit@1          & Hit@3          & Hit@10           \\ 
    \hline
    \multirow{3}{*}{BBFS}    & CHAIN     & 24.32          & 18.38          & 25.97          & 36.65               & 39.29          & 37.94          & 40.08          & 42.02            \\
                             & TREE      & 27.32   & 21.05  & 29.60   & 39.39       & 40.04   & 38.77  & 41.00   & 42.41     \\ 
    &$\Delta$ & +3.00 &+2.67 &+3.63 &+2.74 &+0.75 &+0.83 &+0.92 &+0.39 \\
    \hline
    \multirow{3}{*}{AMIE}    & CHAIN     & 22.60          & 17.25          & 24.27          & 33.78             & 36.21          & 36.06          & 36.31          & 36.47            \\
                             & TREE      & 25.70  & 20.20  & 27.93  & 36.56   & 36.24  & 36.08  & 36.37  & 36.50    \\ 
    &$\Delta$ &+3.10 &+2.95 &+3.66 &+2.78 &+0.03 &+0.02 &+0.06 &+0.03 \\
    \hline
    \multirow{3}{*}{AnyBurl} & CHAIN     & 32.74          & 23.94          & 35.75          & 50.98             & 48.42          & 44.22          & 50.99          & 56.03            \\
                             & TREE      & 35.05  & 26.52  & 38.34  & 52.42   & 48.98  & 45.27  & 51.16 & 55.83   \\
    &$\Delta$ &+2.31 &+2.58 &+2.59 &+1.44 &+0.56 &+1.05 &+0.17 &-0.20 \\
    \hline
    \multicolumn{1}{l}{Dataset}          & \multicolumn{5}{c}{UMLS}                        & \multicolumn{5}{c}{YAGO3-10}          \\ 
    \hline
    \multicolumn{2}{l}{Rule}             & MRR            & Hit@1          & Hit@3          & Hit@10            & MRR            & Hit@1          & Hit@3          & Hit@10           \\ 
    \hline
    \multirow{3}{*}{BBFS}    & CHAIN      & 75.13          & 65.17          & 82.29          & 91.33      & 53.47          & 47.56          & 58.34          & 63.32            \\
                             & TREE      & 77.84  & 68.20   & 85.45  & 92.98     & 54.68   & 49.18  & 59.42   & 63.72     \\ 
    &$\Delta$ & +2.71 &+3.03 &+3.16 &+1.65  &+1.21 &+1.62 &+1.08 &+0.40 \\
    \hline
    \multirow{3}{*}{AMIE}    & CHAIN     & 39.17          & 32.16          & 45.10          & 50.83              & 52.07          & 46.68          & 57.08          & 60.74         \\
                             & TREE     & 42.08  & 35.94  & 46.97  & 51.90  & 53.02  & 48.08  & 57.62  & 60.74  \\ 
    &$\Delta$ &+2.91 &+3.78 &+1.87 &+1.07 &+0.95 &+1.40 &+0.54 &+0.00 \\
    \hline
    \multirow{3}{*}{AnyBurl} & CHAIN    & 69.64          & 55.85          & 79.60          & 92.25             & 63.07          & 57.34          & 67.30          & 72.10        \\
                             & TREE     & 77.43 & 66.73 & 85.94  & 94.67  & 63.38  & 57.82  & 67.36  & 72.38  \\
    &$\Delta$ &+7.79  &+10.88 &+6.34 &+2.42 &+0.06 &+0.48 &+0.06 &+0.28  \\
    \hline

\end{tabular}
\caption{The link prediction performance of different rules on four KGs. CHAIN denotes the original chain-like rules mined by each method. TREE denotes the refined tree-like rules. $\Delta$ denotes the improvements.}
\label{main}
\end{center}
\end{table*}

\begin{table}[t]
\begin{center}
\begin{adjustbox}{width=\columnwidth}
\begin{tabular}{lllll}
    \hline
    KG &  FB15k-237 & WN18RR & UMLS & YAGO3-10 \\
    \hline
    density & 2.59e-03 & 1.06e-04 & 2.20e-01 & 1.42e-04 \\
    avg$\Delta$MRR & 2.80 & 0.61 & 4.47 & 0.74\\
    \hline
\end{tabular}
\end{adjustbox}
\caption{The edge density and the average $\Delta$MRR brought by tree-like rules of each KG. }
\label{density}
\vspace{-0.5cm}
\end{center}
\end{table}

\subsection{Datasets}
\noindent
We employ four commonly used Knowledge Graphs and their corresponding link prediction benchmarks: FB15k-237 \cite{FB15k-237}, WN18RR \cite{ConvKGE}, UMLS \cite{umls}, YAGO3-10 \cite{suchanek2007yago} for the evaluations.

\subsection{Chain-like Rules}
\noindent
We adopt the following methods to mine chain-like rules to serve as the original chain-like rules for evaluation and refinement:
\begin{itemize}[itemsep=1pt,topsep=1pt,parsep=0pt,leftmargin=*]
    \item \textbf{BBFS} We propose a bi-directional breadth-first search method to mine all chain-like rules within length $n$ in KG as a basic search-based rule induction method. 
    \item \textbf{AMIE} \cite{amie} AMIE is a classic symbol-based rule mining system. It learns chain-like rules by adding dangling atoms to the rule body sequentially while evaluating their coverage and confidence.
    \item \textbf{AnyBurl} AnyBurl \cite{anyburl} is a novel RL-based rule induction method and is currently one of the best symbolic rule reasoning methods competing with SOTA embedding reasoning approaches.
\end{itemize}


\subsection{Implementations}
For the tree-like rule refinement process, as it involves many multiplications of large and sparse matrices, we adopt the \emph{torch.sparse} library to help us store and operate such matrices.
For the evaluation process, to fairly evaluate the chain-like rules mined by each method and the tree-like rules refined by us, we adopt the toolkit of AnyBurl \cite{anyburl} to apply the learned rules to KGs and evaluate the link prediction results with the metrics MRR and Hit@n. 
As AnyBurl only originally supported chain-like rules, we modified the AnyBurl toolkit to make it compatible with tree-like rules. 

\subsection{Setups}
\noindent
For each rule, we sample $b = 100$ entities to ground the query variable $X$ and conduct the forward \& backward reasoning process.
We set $\beta$ to the Standard Confidence $sc$ of the original chain-like rule $\mathscr{R}$. 
For each variable, branch atoms with top $k = 5$ scores are selected to refine the rule. The chain-like rules are all within length $n = 3$. The random seed is fixed to 37.


\subsection{Standard Confidence}
\label{sec:sc}
We first verify if the refined tree-like rules are actually more precise, i.e. have higher Standard Confidence than the original chain-like rules. As shown in Table~\ref{tab:sc}, The refined tree-like rules (TREE) consistently have significant higher average Standard Confidence than the original chain-like rules (CHAIN). These results verify that our proposed refinement method effectively refine optimize the initial objective in Eq~\ref{eq:base_objective}. Moreover, we can observe that among the tree types of proposed candidate branch atoms, the ranking of their Standard Confidence is ENT > QRY > AUX, indicating that their constraining strength as branch atoms weakens in this order, which aligns with our intuition.

\subsection{Link Prediction}
\noindent
\label{sec:lp}
To further verify the effective of the refined tree-like rules in the KG reasoning, we evaluated the link prediction performances of chain-like rules from all induction methods and their refined tree-like rules on all four different KGs.
As shown in Table~\ref{main}, the refined tree-like rules consistently outperform the original chain-like rules induced by different methods on different KGs. 
On the FB15k-237 and UMLS datasets, the refinement of chain-like rules into tree-like rules exhibits a performance gain of more than 2\% in most cases. Notably, on the UMLS dataset, tree-like rules demonstrate a significant out-performance compared to Anyburl chain-like rules, with an impressive 7.79\% improvement in MRR. These results lead us to the conclusion that our framework effectively refines chain-like rules from different rule-mining methods into higher-quality tree-like rules on different knowledge graphs, thereby yielding superior reasoning outcomes.

\subsection{Performance Analysis}
\noindent
From Table~\ref{main}, we can also observe that the improvements (delta values) vary across different knowledge graphs.
We explain this based on the topological structure and density of different knowledge graphs.
Intuitively, the sparser the knowledge graph, the simpler the abstract structure it implies (tending towards simple chain-like rules), while the denser the knowledge graph, the more complex the abstract structure it implies (tree-like rules will have an advantage in reasoning).

From a qualitative perspective, WN18RR and YAGO3-10, with smaller deltas, are subsets of WN18 (which is also a subset of WordNet \cite{fellbaum2010wordnet}) and YAGO3, respectively. FB15k-237, with a moderate delta, is a larger subset of FB15k and is relatively denser. These three knowledge graphs have suffered varying degrees of information loss as they are sampled from the whole knowledge graphs. UMLS, on the other hand, is an unfiltered and complete knowledge graph, and therefore retains the most complete information.

From a quantitative perspective, we calculated the density of the four knowledge graphs used in the experiments using edge density (number of edges / the number of possible edges) and the average $\Delta$MRR brought by tree-like rules, and the results are shown in Table~\ref{density}.
We can see that in knowledge graphs with higher density, the reasoning gain brought by tree-like rules is greater, and the Pearson correlation coefficient between them is 0.844. 
These verify that the density of the used knowledge graph and the improvement brought by tree-like rules are positively correlated.

\subsection{Case Study}
To better compare and present the tree-like rules refined from chain-like rules, we provide two specific examples from YAGO3-10 in Figure~\ref{fig:rule_refinement_demo}. It can be observed that the original chain-like rules, constrained by the semantic expressiveness of their chain structure, tend to produce a large number of factually incorrect groundings, resulting in lower standard confidence. In contrast, the tree-like rules refined by our method address the semantic gaps of the original rules in various ways, thereby achieving higher standard confidence.

\begin{figure*}[t]
\begin{tcolorbox}[colframe=blue!50!black, colback=blue!2, title=Examples of Rule Refinement]
\textbf{Chain-like Rule 1} \\
\textbf{Standard Confidence:} 0.39 \\
\texttt{isLocatedIn(X, Y) <= hasCapital(X, Y)}

\textbf{Refined Tree-like Rules:}
\begin{enumerate}
    \item \textbf{Standard Confidence:} 1.00 \\
    \texttt{isLocatedIn(X, Y) <= hasCapital(X, Y), hasOfficialLanguage(Y, M)}
    \item \textbf{Standard Confidence:} 0.50 \\
    \texttt{isLocatedIn(X, Y) <= hasCapital(X, Y), hasWonPrize(Y, M)}
    \item \textbf{Standard Confidence:} 1.00 \\
    \texttt{isLocatedIn(X, Y) <= hasCapital(X, Y), is(Gangtok, Y)}
\end{enumerate}

\textbf{Chain-like Rule 2} \\
\textbf{Standard Confidence:} 0.11 \\
\texttt{worksAt(X, Y) <= hasAcademicAdvisor(X, A), graduatedFrom(A, Y)}

\textbf{Refined Tree-like Rules:}
\begin{enumerate}
    \item \textbf{Standard Confidence:} 0.17 \\
    \texttt{worksAt(X, Y) <= hasAcademicAdvisor(X, A), graduatedFrom(A, Y), \\      
    \phantom{worksAt(X, Y) <=} influences(A, X)}
    \item \textbf{Standard Confidence:} 0.33 \\
    \texttt{worksAt(X, Y) <= hasAcademicAdvisor(X, A), graduatedFrom(A, Y), \\ 
    \phantom{worksAt(X, Y) <=} is(University\_of\_Cambridge, Y)}
    \item \textbf{Standard Confidence:} 0.20 \\
    \texttt{worksAt(X, Y) <= hasAcademicAdvisor(X, A), graduatedFrom(A, Y), owns(Y, M)}
\end{enumerate}
\end{tcolorbox}
\caption{Examples of the refinement of chain-like rules into tree-like rules from YAGO3-10, along with their respective Standard Confidences.}
\label{fig:rule_refinement_demo}
\end{figure*}


\section{Related Work}
\noindent
Rule induction over knowledge graphs is a classical yet challenging task. Inductive Logic Programming (ILP) seeks to induce the symbolic pattern behind the knowledge graphs. It faces the combinatorial space of searching predicates and binding variables.
Traditional ILP methods including AMIE \cite{amie}, AMIE+ \cite{amie+} and RLvLR \cite{RLvLR} used search-based methods to induce chain-like rules. Recently, many works have studied the way of inducing chain-like rules in a differentiable approach, named Differentiable ILP, like NeuralLP \cite{neuralLP}, NLM \cite{nlm}, DLM \cite{dlm}, NLIL \cite{NLIL}. 

However, only a few works considered the defect of expressive power chain-like rules. NLIL \cite{NLIL} induced conjunctions of chain-like rules like $Car(X) \Leftarrow Wheel(Y_1) \wedge Of(Y_1, X) \wedge Window(Y_2) \wedge Of(Y_2, X)$, but they are only tree-like rules with branches at the target variable $X$. TyRule \cite{TyRule} proposed to learn typed rules with type predicate $\textrm{type}_n(x_i)$ adding to each variable, but it needs extra type information of the knowledge graphs. In this paper, we propose the concept of tree-like rules and an effective framework for refining chain-like rules into tree-like rules to reach better reasoning ability.

\section{Conclusion}
\noindent
This paper introduces the concept of tree-like rules and presents an effective framework for refining chain-like rules into tree-like rules. To verify the effectiveness of the tree-like rules refined by our framework, this paper carried out experiments to show that the refined tree-like rules consistently exhibit higher Standard Confidence and outperform the original chain-like rules on KG reasoning tasks. Further analysis illustrates that the improvements brought by the tree-like rules are positively correlated with the the density of the KGs.

\section{Acknowledgments}
\noindent
This work was supported by the National Key R\&D Program of China (No. 2022ZD0160503) and and Beijing Natural Science Foundation (L243006). This work was also supported by the National Natural Science Foundation of China (No.62376270).

\bibliographystyle{ACM-Reference-Format}
\bibliography{custom}

\appendix


\end{document}